%% file: main.tex
\documentclass{article} 
\usepackage{iclr2024_conference,times}

\input{math_commands.tex}

\usepackage{hyperref}
\usepackage{url}
\usepackage{authblk}
\usepackage{amssymb}
\usepackage{enumitem}
\usepackage{graphicx}
\usepackage{booktabs}
\usepackage{multirow}
\usepackage{subcaption}
\usepackage{pythonhighlight}
\usepackage{arydshln}

\newcommand*{\affaddr}[1]{#1} 

\title{AgentOhana: Design Unified Data and Training Pipeline for Effective Agent Learning}
\iclrfinalcopy
\author{{Jianguo~Zhang}$^{*}$, 
\text{Tian~Lan\thanks{Equal contributions.}~~}, 
\text{Rithesh Murthy},
\text{Zhiwei Liu},
\text{Weiran Yao},
\text{Ming Zhu},
\text{Juntao Tan},
\text{Thai Hoang},
\text{Zuxin Liu},
\text{Liangwei Yang},
\text{Yihao Feng},
\text{Shirley Kokane},
\text{Tulika  Awalgaonkar},
\text{Juan Carlos Niebles},
\text{Silvio Savarese},
\text{Shelby Heinecke},
\text{Huan Wang},
\text{Caiming Xiong}\\
\affaddr{Salesforce Research, USA}\\
\affaddr{jianguozhang@salesforce.com, tian.lan@salesforce.com}
}

\begin{document}

\maketitle

\begin{abstract}

Autonomous agents powered by large language models (LLMs) have garnered significant research attention. However, fully harnessing the potential of LLMs for agent-based tasks presents inherent challenges due to the heterogeneous nature of diverse data sources featuring multi-turn trajectories. In this paper, we introduce \textbf{AgentOhana} as a comprehensive solution to address these challenges. \textit{AgentOhana} aggregates agent trajectories from distinct environments, spanning a wide array of scenarios. It meticulously standardizes and unifies these trajectories into a consistent format, streamlining the creation of a generic data loader optimized for agent training. Leveraging the data unification, our training pipeline maintains equilibrium across different data sources and preserves independent randomness across devices during dataset partitioning and model training. Additionally, we present \textbf{xLAM-v0.1}, a large action model tailored for AI agents, which demonstrates exceptional performance across various benchmarks. Begin the exploration at \url{https://github.com/SalesforceAIResearch/xLAM}.
\end{abstract}

\input{01-introduction}

\input{02-methodology}

\input{03-experiments}

\section{Conclusion}
In conclusion, the creation of \textit{AgentOhana} represents a significant step forward in addressing the challenges inherent in consolidating diverse data of the multi-turn LLM agent trajectories. Through the development of unified data and training pipelines, we have established a framework capable of handling the intricacies of various data structures and formats, thereby ensuring compatibility across a multitude of environments. By providing a comprehensive and high-quality dataset, we aim to empower researchers and practitioners to push the boundaries of AI capabilities, ultimately contributing to the advancement of autonomous agents powered by LLMs.

\newpage
\bibliography{reference_agent}
\bibliographystyle{iclr2024_conference}

\clearpage
\appendix

\begin{center}
\Large
\textbf{Appendix}
\end{center}
\input{04-appendix}

\end{document}

%% file: math_commands.tex
\usepackage{amsmath,amsfonts,bm}

\def\eqref#1{equation~\ref{#1}}

\def\1{\bm{1}}

\DeclareMathAlphabet{\mathsfit}{\encodingdefault}{\sfdefault}{m}{sl}
\SetMathAlphabet{\mathsfit}{bold}{\encodingdefault}{\sfdefault}{bx}{n}



%% file: 01-introduction.tex
\section{Introduction}

Large language models (LLMs) have shown strong abilities in code generation, mathematical reasoning, conversational AI, and AI agents~\citep{openai2023gpt4,jiang2023mistral,zhang2023dialogstudio,liu2023agentbench,nijkamp2023codegen2}.  Among these, LLM-powered autonomous agents are gaining increasing attention. Recent frameworks for LLM agents, such as AutoGPT~\citep{autogpt23}, OpenAgent~\citep{xie2023openagents}, BOLAA~\citep{liu2023bolaa}, XAgent~\citep{xagent2023}, and LangChain~\citep{langchain23}, have been designed to support agent tasks, and they have attracted significant interest in the open-source community.

Nevertheless, many existing agents are powered by closed-source LLM APIs such as GPT-4~\citep{openai2023gpt4} and Gemini~\citep{team2023gemini}, mainly because most open-source models struggle to perform long-horizon reasoning and handle complex agent tasks~\citep{liu2023agentbench, liu2023bolaa}. Recently, there have been ongoing efforts on training open-source models instead of relying solely on commercialized APIs. For instance, AgentLM~\citep{zeng2023agenttuning}, Lemur~\citep{xu2023lemur} and Lumos~\citep{yin2023lumos} are trained for agents based on Llama-2 family~\citep{touvron2023llama}, along with special reasoning, planning and acting prompts design such as React~\citep{yao2023react}, Self-Reflection~\citep{shinn2023reflexion, madaan2023self, wang2023drdt} to enhance the abilities.
On the same, there are works to open-source agent relevant data and train agent models such as ToolLlama~\citep{qin2023toolllm}, ToolAlpaca~\citep{tang2023toolalpaca} and API-bank~\citep{li2023api} to enhance abilities on reasoning, tool usages and plannings. They have shown impressive performance on agent relevant tasks. 

However, navigating the data landscape for LLM agents becomes increasingly intricate when dealing with non-standardized data formats sourced from diverse dataset collections, especially those featuring interactions of multi-turns, as commonly observed in agent-relevant data.
The heterogeneity in data structures, syntaxes, labeling conventions, and processing methods across datasets poses a formidable challenge, complicating the training and fine-tuning processes of LLMs. 
The lack of standardized formats introduces complexities in harmonizing disparate data sources, leading to potential biases and inconsistencies. 
Addressing these challenges requires developing robust preprocessing pipelines, ensuring unification and compatibility across varied data formats, and implementing strategies to mitigate biases that may arise from non-standardized representations. 
With the increasing demand for comprehensive and diverse datasets, establishing effective methods to manage non-standardized data formats is crucial for ensuring the robust performance of LLM agents across a spectrum of applications.

In this work, we bridge the existing gap by building the first comprehensive agent data collection and training pipeline, named \textit{AgentOhana}. 
Drawing inspiration from the notable achievements of DialogStudio~\citep{zhang2023dialogstudio} and FLAN~\citep{longpre2023flan} in the realms of Conversational AI and instruction-based fine-tuning, 
\textit{AgentOhana} is tailored to accommodate the wide variety of data structures and formats encountered in LLM agent trajectories. It employs specialized processes to transform disparate data into a uniform format, achieving seamless integration across multiple sources.
Furthermore, the collection undergoes a meticulous filtering process to ensure high-quality trajectories, 
thereby introducing an extra layer of quality control. 
Leveraging the data standardization and unification, our training pipeline preserves independent randomness across devices during both dataset partitioning and model training, thus avoiding the inadvertent introduction of biases into the training process.
This comprehensive approach guarantees that AgentOhana not only unifies trajectories across environments but also enhances the overall quality and the reliability of the collected data, as well as the performance and the robustness of the model. Our approach ensures that AgentOhana serves as a versatile and accessible resource for the research community, streamlining the development process for future applications.
The contributions of this paper are as follows: 

\begin{itemize}[leftmargin=*]
    \item \textbf{Innovative Solution to Data Heterogeneity}: We introduce \textit{AgentOhana}, a pioneering platform designed to address the complex challenges associated with the consolidation of heterogeneous data sources pertaining to multi-turn LLM agent trajectories. This contribution represents a critical step forward in overcoming the obstacles of data diversity and fragmentation.
    \item \textbf{Extensive Environmental Coverage}: \textit{AgentOhana} distinguishes itself by incorporating agent data from ten distinct environments, spanning a comprehensive array of scenarios. This diverse collection facilitates a broad spectrum of research opportunities, enabling investigations into various aspects of agent behavior and interaction.
    \item \textbf{Data Standardization and Unification}: A core achievement of this work is the systematic standardization and unification of LLM agent data into a consistent format. This process has enabled the creation of a generic data loader, optimizing the dataset for agent training that maintains equilibrium across different data sources and preserves independent randomness across devices. 
    \item \textbf{Large Agent Model}: We have developed XLAM-v0.1, a robust large action model tailored for AI agents. Demonstrating strong performance across three rigorous benchmarks, XLAM-v0.1 showcases the potential of \textit{AgentOhana} in facilitating the training of high-performing AI agents.
\end{itemize}

\begin{figure}[h]
    \centering
    \includegraphics[width=0.95\linewidth]{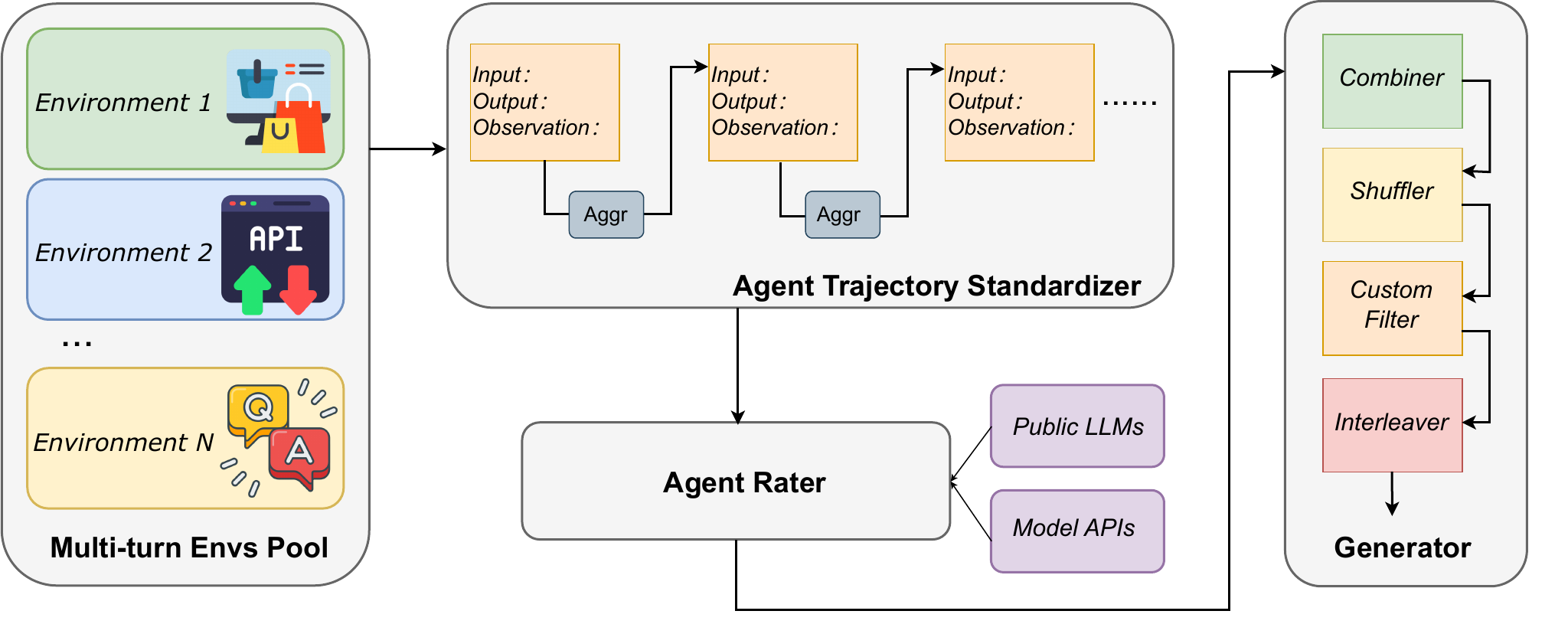}
    \caption{Workflow of AgentOhana. A homogeneous multi-turn data format is designed to consolidate heterogeneous trajectories from diverse data sources. \emph{AgentRater} then assesses and filters agent trajectories. Finally, a streaming data loader enables integration of various datasets and feeds data into a distributed training process at random.}
    \label{fig:unified_dataset}
\end{figure} 





%% file: 02-methodology.tex
\section{Methodology}

As illustrated in the workflow of \emph{AgentOhana} shown in Figure \ref{fig:unified_dataset}, we adopt a homogeneous multi-turn data format designed to consolidate trajectories from heterogeneous data sources. Additionally, we introduce a method called \emph{AgentRater} to assess and filter agent trajectories based on public or close-world models. Finally, we adopt a generic dataloader as a central component to enable smooth integration of various datasets into a distributed training process.

\begin{figure}[t!]
    \centering
    \includegraphics[width=1.0\linewidth]{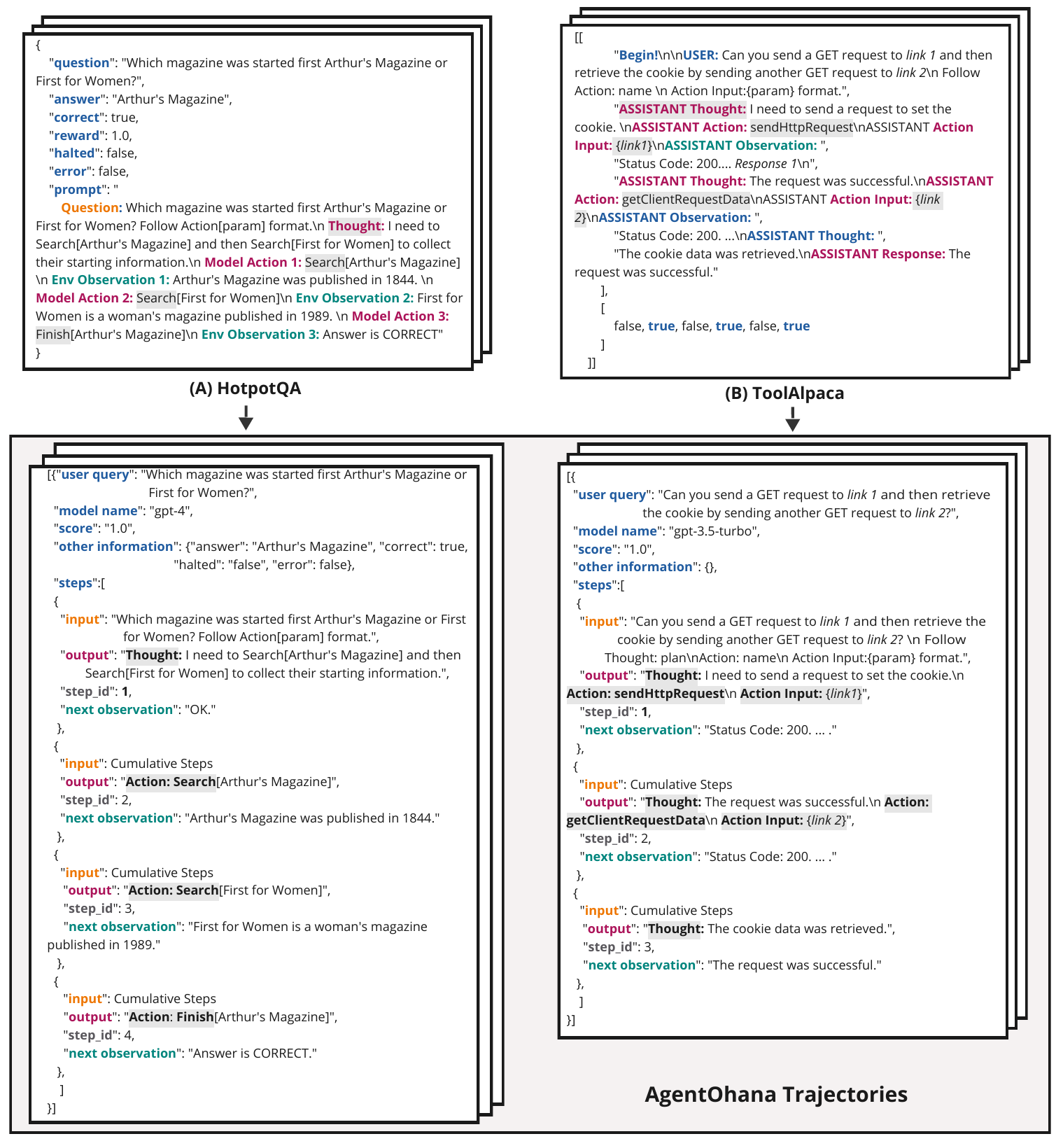}
    \caption{Example trajectories from (A) HotpotQA, (B) ToolAlpaca to AgentOhana.}
    \label{fig:example-trajectory}
\end{figure}

\subsection{Heterogeneity of various datasets}


The formats of agent data vary significantly across different environments, posing significant difficulties and challenges in unifying data, training, and analyzing models. As illustrated in row 1 of Figure \ref{fig:example-trajectory},, trajectories from two distinct environments show markedly different data organization methods, a phenomenon observed universally across different environments.
For instance, the HotpotQA environment~\citep{liu2023bolaa} consolidates the whole target trajectory into a single string under the \textit{prompt} key. This design  requires efforts to retrieve \textit{user query}, \textit{Thought}, \textit{Model Action: i} along with \textit{Env Observation: i} for each step  $i^{th}\in [1,N]$ from a single string. Conversely, ToolAlpaca requires the identification and matching of prompt inputs, model outputs, and observations at each step, followed by the accurate aggregation of trajectory history prior to proceeding to the next step. Appendix \ref{fig:appendix-raw-trajectory} shows more examples of the original trajectories from four environments.


\subsection{Homogeneous multi-turn agent trajectory standardization}


To address the challenges identified,  we propose a unified agent data format, as depicted in row 2 of Figure \ref{fig:example-trajectory}, showcasing our proposed unified trajectory data format.  We construct a homogeneous JSON dictionary format to encapsulate all relevant content of each trajectory. Concretely, our format incorporates all important elements such as \textit{user query} to store the initial user query, \textit{model name} to identify the corresponding model  and \textit{score} to log the available model performance score. These elements can be used to differentiate between models and are poised to facilitate the development of pairwise samples for cutting-edge training methodologies such as DPO~\citep{rafailov2023direct}, self-reward~\citep{yuan2024self} and AI feedback~\citep{guo2024direct} LLMs. Additionally, we save auxiliary  trajectory information or specific notes into \textit{other information}, providing a  reference for further analysis or model improvement initiatives.


To enhance the preservation and analysis of multi-turn agent trajectory information, we propose a structured definition of a \textit{step} that captures the details of each interaction turn. A step comprises three main components: \textit{input}, \textit{output}, and \textit{next observation}. The \textit{input} component consolidates the current prompt and a historical record of past interactions, serving as a comprehensive context for the interaction. The \textit{output} component captures the model's predictions, detailing its decision-making and planning. The \textit{next observation} component records the environment's feedback, essential for the feedback loop and system adaption.

Our framework employs a predefined method for aggregating interaction history within the \textit{input} component, effectively concatenating inputs and outputs from previous steps to construct a comprehensive context. Specifically, at the $i^{th}$ step, the input is formatted as  \emph{input of step 1, Action: output of step 1, Observation: next observation of step 1, ..., input of step i-1, Action: output of step i-1, Observation: next observation of step i-1}.  This approach ensures a detailed chronological account of interactions, facilitating a nuanced understanding of the trajectory.

While this default aggregation strategy of \textit{input} is integral to our framework, we also accommodate the customization of data compilation methods. Users are encouraged to explore alternative strategies that exploit the structured \textit{input}, \textit{output}, and \textit{next observation} components, tailoring the data format to their specific research or application needs. Figure \ref{fig:example-trajectory}, row 2, illustrates the transformation of trajectories from environments such as HotpotQA and ToolAlpaca using our defined step structure, where \textit{output} aligns with the format specifications in the initial prompt input, demonstrating the framework's adaptability and practical utility.

By standardizing the capture of interactions between agents and their environments, this methodology not only facilitates a uniform approach to data documentation but also enhances the potential for in-depth analysis and refinement of AI models. This is achieved by providing a granular view of the agent interactions, decision-making process and its results, thereby enabling a more nuanced understanding and improvement of model performance.



\label{sec:standard_raw_data}

\subsection{AgentRater}
\label{sec:agent_rater}

\begin{figure}[t!]
    \centering
    \includegraphics[width=1.0\linewidth]{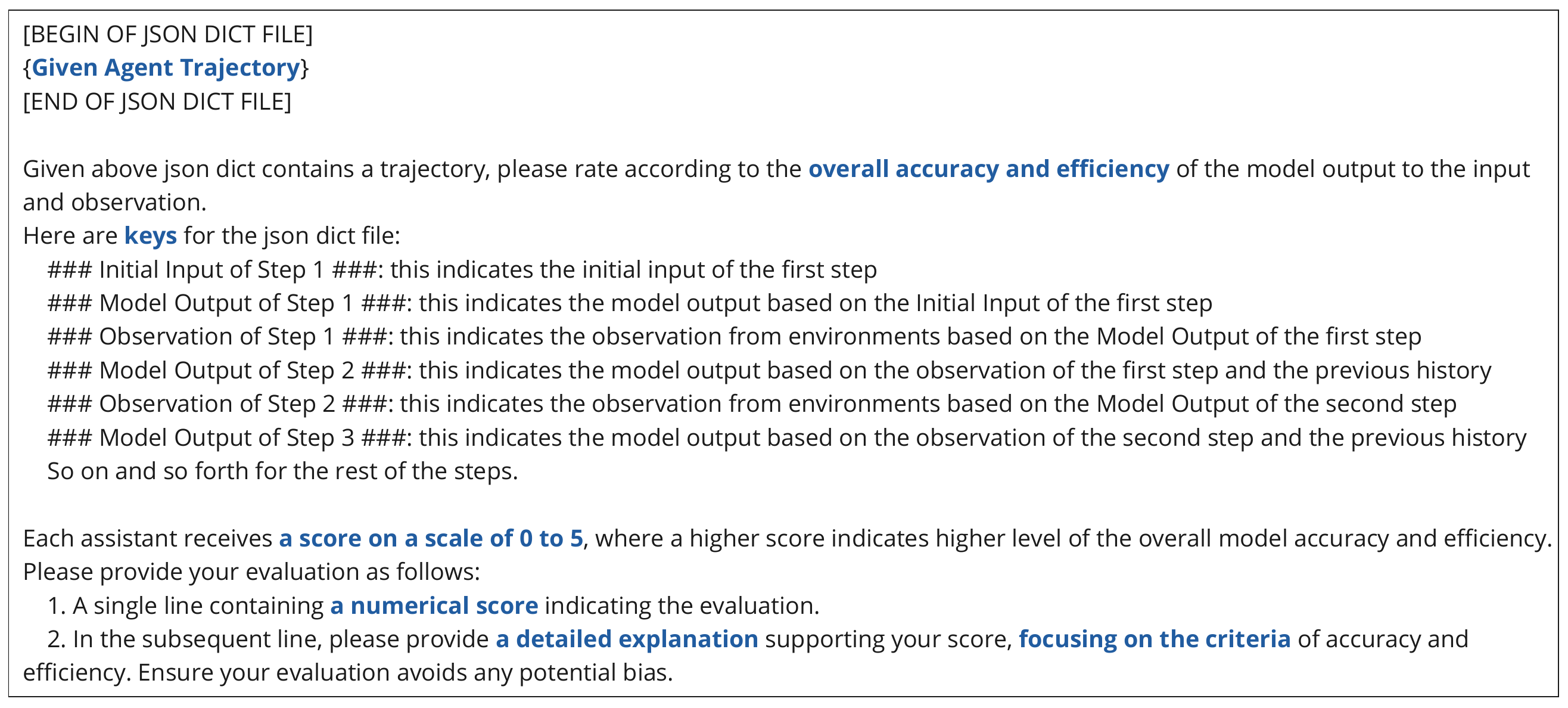}
    \caption{A prompt template for the AgentRater, where an open-source model (e.g., Mistral) or close-world API (e.g., ChatGPT) will rate the whole agent trajectory based on criterias and then assign a score from 0 - 5.}
    \label{fig:rate-prompt}
\end{figure}

Agent trajectories represent a complex subset of data distinct from general and straightforward instructional data. While datasets like Alpaca~\citep{dubois2023alpacafarm} features single-turn examples, and LMSYS-Chat~\citep{zheng2023lmsys} includes dialogues averaging around two turns, these generally encompass simpler interaction patterns. DialogStudio~\citep{zhang2023dialogstudio} does offer multi-turn dialogue examples, these are primarily confined to conversations between the user and the system, lacking interactions with external environments.


In contrast, agent trajectories delve into more intricate scenarios where an agent interacts with complex environments such as websites, APIs, and tools. This complexity is heightened by the agent's capacity to communicate with other agents, navigate through diverse interfaces, and undertake tasks that require a sequence of interactions rather than single or limited exchanges.

The challenge with agent trajectories extends to the evaluation of performance and quality. While some environments offer rewards as feedback for an agent's trajectory, such rewards are often tied to the task's final outcome rather than reflecting the quality of the trajectory itself. Consequently, a high reward does not necessarily indicate a flawless trajectory. For example, an agent might generate invalid actions during intermediate steps of a task. Here is partial trajectory from the Webshop environment~\citep{yao2022webshop}: \emph{model output of step 4: "click[old world style]", observation of step 4: "You have clicked old world style."; model output of step 5: "click[rope sausage]", observation of step 5: "Invalid action!"; model output of step 6: "", observation of step 6: "Invalid action!"; model output of step 7: "click[Buy Now]", observation of step 7: "Your score (min 0.0, max 1.0): 1.0"}, where the agent randomly clicks other buttons in Webshop website or generates empty responses before buying an item. 


\begin{table}[]
\centering
\begin{tabular}{lccc}
\hline
\multicolumn{1}{c}{\textbf{Environments \& Data}} & \multicolumn{1}{c}{\textbf{\#Sampled Trajs}} & \multicolumn{1}{c}{\textbf{\#Filtered Trajs}} & \multicolumn{1}{c}{\textbf{\#Average Turns}} \\ \hline
Webshop~\citep{yao2022webshop}               &     11200   & 2063    &    6.8      \\
AlfWorld~\citep{ALFWorld20}         &  954     &      336            &      13.5                \\
HotpotQA~\citep{yang2018hotpotqa}      & 1740   &        402       &          4.8                 \\
ToolBench~\citep{qin2023toolllm}            &  83771  &           30319     &    3.1         \\
ToolAlpaca~\citep{tang2023toolalpaca}        &       3936    &     3399    &       2.5           \\
Operating System~\citep{liu2023agentbench}       &   647    &        195          &         3.9     \\
APIbank~\citep{li2023api}         &    33415     &  4902        &          1.0         \\
DataBase~\citep{liu2023agentbench}               &    6376        &      538    &     2.0      \\
Mind2Web~\citep{deng2023mind2web}                &      23378       &            122         &    1.0                                  \\
Knowledge Graph~\citep{liu2023agentbench}        &     2501    &    324       &     6.0    \\ \hdashline

AgentOhana  &     167918    &    42600      &     3.1    \\ \hline

\end{tabular}\caption{Statistics of AgentOhana. AgentOhana consists of data from 10 different environments. \textit{\#Sampled Trajs} indicates the trajectories sampled and filtered from the original environment,  \textit{\#Filtered Trajs} indicates the filter trajectories with the AgentRater score $>=4.0$, \textit{\#Average Turns} indicates average number of turns in the filtered trajectories. Among the environments, \textit{Operating System}, \textit{DataBase}, \textit{Mind2Web} and \textit{Knowledge Graph} are derived from ~\citep{zeng2023agenttuning}. Additionally, AgentOhana integrates partial general instruction data sourced from DialogStudio~\citep{zhang2023dialogstudio}, this subset is not represented in the table.}
\end{table}\label{table: data-stats}

To mitigate the issues, we design a method, named \textit{AgentRater} to rate the agent trajectory based on strong public models such as Mistral~\citep{jiang2023mistral} or close-world APIs such as ChatGPT~\citep{openai2023gpt4}. Different with approaches~\citep{zhang2023dialogstudio,chen2023alpagasus} where they rate each triplet of (instruction, input, response) pair on general instruction data as there are usually single-turn or short conversations, we rate the whole trajectory on agent data.  Figure \ref{fig:rate-prompt} shows a corresponding prompt template, where we rate the trajectory with a score 0-5 and an explanation, and they can be used to further develop better AgentRater models. Table \ref{table: data-stats} shows the statistics of AgentOhana.

\subsection{Generic Dataloader}
As the protocol involves loading data in the correct format for the trainer, the implementation of a generic dataloader becomes crucial in harmonizing the entire data and training pipeline. This dataloader serves as a central component, facilitating seamless integration of diverse datasets into the training process. Its generic nature ensures flexibility and compatibility across various data formats, enabling efficient data ingestion before feeding into the training framework. 

\subsubsection{AgentModelDatasetBase}

We have introduced the \emph{AgentModelDatasetBase} class to streamline common tasks such as prompt formatting while providing a virtual template for creating individual datasets. 
While loading data may appear straightforward at this stage, there are still several intricate issues to address. 
For instance, in addition to employing a machine-assisted filter as detailed in Section \ref{sec:agent_rater}, users may prefer a certain level of control over data quality.
Moreover, the randomness of data batching from different datasets could pose challenges, particularly when dealing with distributed training among multiple devices, which requires a 
a comprehensive approach to
ensure robust dataset management. 

\subsubsection{Custom Datasets Creation}
\paragraph{Individual dataset}
As depicted in the following example, we begin by loading individual raw data prepared from Section \ref{sec:standard_raw_data}, typically via the streaming mode. Then, for each dataset, we can optionally introduce the filter generator to further customize the selection of data just before feeding it into the trainer. For instance, in the following example, data with relatively low scores will be further evaluated and removed. Finally, we shuffle this dataset randomly with controlled seeding to ensure randomness and reproducibility.

\begin{minipage}{\linewidth}
\begin{python}
class WebshopMultiTurn(AgentModelDatasetBase):
    
    # we can further filter out trajectories at this stage
    @staticmethod
    def _high_score_filter_generator(data, score=0.8):
        for d in data:
            if d["score"] >= score:
                yield {"prompt": d["input"], "chosen": d["output"]}

    def create_datasets(self, seed=None):
        train_data = load_dataset(
            ...
            streaming = self.args.streaming,
        )
        train_data = IterableDataset.from_generator(
            self._high_score_filter_generator, 
            gen_kwargs={"data": train_data}
        )
        train_data = train_data.shuffle(seed=seed, buffer_size=1000)
        return train_data
\end{python}
\end{minipage}

\paragraph{Combined Datasets}
Our primary focus in combining datasets lies in ensuring randomness during the batching process, particularly when dealing with multiple datasets. To achieve this, we employ the \emph{init$\_$device$\_$seed} function to diversify the controlled seeds based on the process ID when data parallelism is utilized across multiple devices. By carefully managing the seeding process, we aim to maintain a balanced distribution of data by partitioning, shuffling and interleaving data across devices while preserving randomness, thus enhancing the robustness and reproducibility of our training procedure. 

\begin{minipage}{\linewidth}
\begin{python}
toolbench_multi_turn = ToolBenchMultiTurn(tokenizer, script_args)
webshop_multi_turn = WebshopMultiTurn(tokenizer, script_args)
...

data = [toolbench_multi_turn, webshop_multi_turn, ...]

sample_probs = [0.1, 0.1, ...]

# a device-dependent seeding will be utilized based on the combination of 
# the given default seed and the process ID
seed = init_device_seed(seed=42) 

train_dataset, eval_dataset = \
    interleave_data(
        data_objects=data,
        sample_probs=sample_probs,
        seed=seed)
\end{python}
\end{minipage}

%% file: 03-experiments.tex
\section{Experiments}
\subsection{Training}
We adopted a supervised fine-tuning approach to enhance the performance of our agent model, xLAM-v0.1, which was initially pre-trained on the Mixtral-8x7B-Instruct-v0.1 model \citep{jiang2024mixtral}. To execute this fine-tuning process, we capitalized on the capabilities of AgentOhana. Our fine-tuning procedure was conducted concurrently on 8 Nvidia H100 GPUs, utilizing the QLoRA framework \citep{dettmers2023qlora}. 


Our dataset compilation strategy ensured the creation of a comprehensive training corpus with a balanced distribution of data from various sources. Maintaining this equilibrium across different sources mitigated biases and ensured the robustness of our model across diverse inputs. Furthermore, we meticulously preserved independent randomness across devices during dataset partitioning and model training, preventing the introduction of unintended biases into the training process.

Throughout the fine-tuning process, our model traversed each individual dataset approximately 3 times on average. This multi-epoch training regimen facilitated comprehensive exposure to the dataset, enabling the model to effectively learn intricate patterns present in the training data.

\subsection{Benchmarks}


In the subsequent sections, we will present experimental evaluations conducted across four benchmarks: Webshop~\citep{yao2022webshop}, HotpotQA~\citep{yang2018hotpotqa}, ToolEval~\citep{qin2023toolllm}, and MINT-Bench~\citep{wang2023mint}.

Webshop creates an online shopping environment simulating product purchases, while HotpotQA involves multi-hop question-answering tasks requiring logical reasoning across Wikipedia passages via the Wikipedia API. We adopt BOLAA's framework~\cite{liu2023bolaa}, comprising five single-agent settings and a multi-agent scenario, to evaluate model performance. For the Webshop benchmark, BOLAA comprises 900 user queries, of which we serve 200 as a test subset. For HotpotQA, 300 user questions are sampled into three difficulty levels—easy, medium, and hard—with each category containing 100 questions. These questions are exclusively reserved for model testing to ensure a rigorous evaluation process. We use BOLAA's evaluation metrics, average reward for Webshop and F1 score for HotpotQA, to measure model performance. In Webshop, the reward metric assesses model accuracy based on the attributes overlapping between the purchased and the ground-truth items, while in HotpotQA, it quantifies the accuracy of agent-predicted answers against ground-truth responses.

ToolEval is designed for real-time assessment of functional calling capabilities via RapidAPI, initially utilizing GPT-3.5-Turbo-16k as its evaluator. However, after careful investigation, we found GPT-3.5-Turbo-16k unreliable for assessing complex function calls and tool usage scenarios. Consequently, we switched to GPT-4-0125-preview as our primary evaluator. Due to the real-time nature of these evaluations, APIs may experience downtime or timeouts, leading to inconsistency in model comparisons across different time frames. To address this, all models are evaluated within the same time frame. Our evaluation employs the default depth-first search-based decision tree methodology, augmented by the Pass Rate metric to assess an LLM’s ability to execute instructions, a fundamental criterion for optimal tool usage. We present our findings at the first level of the ToolEval evaluation, focusing on three distinct scenarios: (1) unseen instructions with the same set of tools, (2) unseen tools within previously seen categories, and (3) unseen tools from entirely new categories that have not been seen before.

MINT-Bench  benchmark evaluates LLMs’ ability to solve tasks with multi-turn interactions
by using tools and leveraging natural language feedback. The benchmark focuses on
reasoning, coding, and decision-making through a diverse set of established evaluation datasets, and carefully curate them into a compact subset for efficient evaluation.
The benchmark asks LLMs to solve tasks with different interaction limits from 1 to 5 step and quantify LLMs’ tool-augmented task-solving
capability by absolute performance success rate, which measures the percentage
of successful task instances as a function of interaction steps.

\begin{table}[t]
\resizebox{1.0\linewidth}{!}{
\centering
\begin{tabular}{l|ccccccc}
\toprule
\multirow{2}{*}{LLM} &  \multicolumn{6}{c}{LAA Architecture}                                                          \\ \cmidrule(l){2-7} 
                            & ZS     & ZST    & ReAct           & PlanAct               & PlanReAct       & BOLAA           \\ \hline
{Llama-2-70b-chat~\citep{touvron2023llama}}                           & 0.0089 & 0.0102  & 0.4273      & 0.2809                & 0.3966           & 0.4986  \\
{Vicuna-33b~\citep{zheng2023judging}}         & 0.1527 & 0.2122  & 0.1971      & 0.3766                & 0.4032       & 0.5618  \\
{Mixtral-8x7B-Instruct-v0.1~\citep{jiang2024mixtral}}            & 0.4634 & 0.4592 & \underline{0.5638}          & 0.4738              & 0.3339          & 0.5342 \\
{GPT-3.5-Turbo}   & 0.4851  & \underline{0.5058} & 0.5047    & 0.4930      & \underline{0.5436}       & \underline{0.6354}  \\
{GPT-3.5-Turbo-Instruct} & 0.3785 & 0.4195 & 0.4377    & 0.3604    & 0.4851   & 0.5811 \\  
{GPT-4-0613}   & \underline{0.5002} & 0.4783 & 0.4616   & \textbf{0.7950}     & 0.4635     & 0.6129 \\ \hdashline
{xLAM-v0.1}  & \textbf{0.5201} & \textbf{0.5268} & \textbf{0.6486} & \underline{0.6573}       & \textbf{0.6611}      & \textbf{0.6556}          \\
\bottomrule
\end{tabular}}\caption{Average reward on the WebShop environment. \textbf{Bold} and \underline{Underline} results denote the best result and the second best result for each setting, respectively. }
\label{tab:webshop_reward}
\end{table}





\subsection{Webshop}
Table \ref{tab:webshop_reward} showcases the performance of our model within the Webshop environment. xLAM-v0.1 consistently outperforms both GPT-3.5-Turbo and GPT-3.5-Turbo-Instruct across all agent configurations. Moreover, it surpasses GPT-4-0613 in five out of six settings, with the latter demonstrating superior planning capabilities but lower performance in reasoning, self-reflection, and multi-agent interactions. These findings underscore the robust and versatile capabilities of the xLAM model across a variety of agent scenarios.

\begin{table}[t]
\resizebox{1.0\linewidth}{!}{
\centering
\begin{tabular}{l|ccccccc}
\toprule
\multirow{2}{*}{LLM} &  \multicolumn{6}{c}{LAA Architecture}                                                          \\ \cmidrule(l){2-7} 
                            & ZS     & ZST    & ReAct           & PlanAct               & PlanReAct        \\ \hline

{Mixtral-8x7B-Instruct-v0.1~\citep{jiang2024mixtral}}  & 0.3912 & 0.3971 & 0.3714  & 0.3195  & 0.3039 \\
{GPT-3.5-Turbo}   & 0.4196  & 0.3937 & 0.3868    & 0.4182      & 0.3960        \\ 
{GPT-4-0613}   & \textbf{0.5801} & \textbf{0.5709} & \textbf{0.6129}   & \textbf{0.5778} & \textbf{0.5716} \\     \hdashline
{xLAM-v0.1}  & \underline{0.5492} & \underline{0.4776} & \underline{0.5020} & \underline{0.5583}      & \underline{0.5030}         \\
\bottomrule
\end{tabular}}\caption{Average reward on the HotpotQA environment. \textbf{Bold} and \underline{Underline} results denote the best result and the second best result for each setting, respectively. }
\label{tab:hotpotqa_reward}
\end{table}

\subsection{HotpotQA}

Table \ref{tab:hotpotqa_reward} details the results in the HotpotQA environment, highlighting xLAM's superior performance relative to GPT-3.5-Turbo and Mixtral-8x7B-Instruct-v0.1 across all settings. While GPT-4-0613 exhibits a slight performance edge, our analysis on models' predictions reveals that it typically identifies correct answers within four steps, suggesting that it may have been trained on a substantial corpus of relevant question-answering examples, thereby possessing enhanced domain-specific knowledge compared to its counterparts.


\begin{table}[ht]
\resizebox{1.01\linewidth}{!}{
\centering
\begin{tabular}{l|cccc}
\toprule
\hline
    & \begin{tabular}[c]{@{}c@{}}Unseen Insts \& Same Set \end{tabular} & \begin{tabular}[c]{@{}c@{}}Unseen Tools \& Seen Cat \end{tabular} & \begin{tabular}[c]{@{}c@{}}Unseen Tools \& Unseen Cat\end{tabular}     \\ \hline
TooLlama V2~\citep{qin2023toolllm}                   & 0.4385    & 0.4300    & 0.4350                  \\
GPT-3.5-Turbo-0125  & 0.5000    & 0.5150    & 0.4900                 \\
GPT-4-0125-preview              & \textbf{0.5462}  & \underline{0.5450}  & \underline{0.5050}           \\ \hdashline
xLAM-v0.1             & \underline{0.5077}   & \textbf{0.5650}  & \textbf{0.5200}                 \\
\bottomrule
\end{tabular}
}
\caption{Pass Rate on ToolEval on three distinct scenarios. 
\textbf{Bold} and \underline{Underline} results denote the best result and the second best result for each setting, respectively. 
}\label{table:tool-eval}
\end{table}

\subsection{ToolEval}

Table \ref{table:tool-eval} displays the results on ToolEval. xLAM-v0.1 surpasses both TooLlama V2 and GPT-3.5-Turbo-0125 across all evaluated scenarios, and outperforms GPT-4-0125-preview in two out of the three settings. This performance indicates xLAM-v0.1's superior capabilities in function calling and handling complex tool usage tasks. We posit that the model's performance could be enhanced further through data augmentation involving a variety of prompts.



\subsection{Mint-Bench}


Table \ref{table:mint-bench} presents the testing results on the challenging and comprehensive MINT-Bench, with baseline comparisons drawn from the official leaderboard\footnote{\url{https://xwang.dev/mint-bench/}}. The xLAM-v0.1 model secures the third rank in this rigorous benchmark, outperforming other agent-based models such as Lemur-70b-Chat-v1 and AgentLM-70b, as well as general large language models (LLMs) including Claude-2 and GPT-3.5-Turbo-0613. These results highlight the exceptional capability of our model to navigate the complexities of multi-turn interactions and task resolution.


\begin{table}[ht]
\centering
\begin{tabular}{l|cccccc}
\toprule
\hline
    & \begin{tabular}[c]{@{}c@{}}1-step \end{tabular} & \begin{tabular}[c]{@{}c@{}}2-step \end{tabular} & \begin{tabular}[c]{@{}c@{}}3-step\end{tabular} & \begin{tabular}[c]{@{}c@{}}4-step\end{tabular} & \begin{tabular}[c]{@{}c@{}}5-step\end{tabular}    \\ \hline
GPT-4-0613                    & nan    & nan    & nan    & nan    & 69.45                \\
Claude-Instant-1              & 12.12  & 32.25  & 39.25  & 44.37  & 45.90                \\
xLAM-v0.1             & 4.10   & 28.50  & 36.01  & 42.66  & 43.96                \\
Claude-2                      & 26.45  & 35.49  & 36.01  & 39.76  & 39.93
          \\
Lemur-70b-Chat-v1~\citep{xu2023lemur}             & 3.75   & 26.96  & 35.67  & 37.54  & 37.03                \\
GPT-3.5-Turbo-0613            & 2.73   & 16.89  & 24.06  & 31.74  & 36.18                \\
AgentLM-70b ~\citep{zeng2023agenttuning}   & 6.48   & 17.75  & 24.91  & 28.16  & 28.67                \\
CodeLlama-34b~\citep{roziere2023code} & 0.17   & 16.21  & 23.04  & 25.94  & 28.16                \\
Llama-2-70b-chat~\citep{touvron2023llama}   & 4.27   & 14.33  & 15.70  & 16.55  & 17.92            \\
\bottomrule
\end{tabular}
\caption{Testing results on MINT-Bench with different interaction limits from 1 to 5 step.  }\label{table:mint-bench}
\vspace{-0.3cm}
\end{table}




%% file: 04-appendix.tex
\section{Heterogeneity of various datasets}

\begin{figure*}[h]
    \centering
    \includegraphics[width=1.1\linewidth]{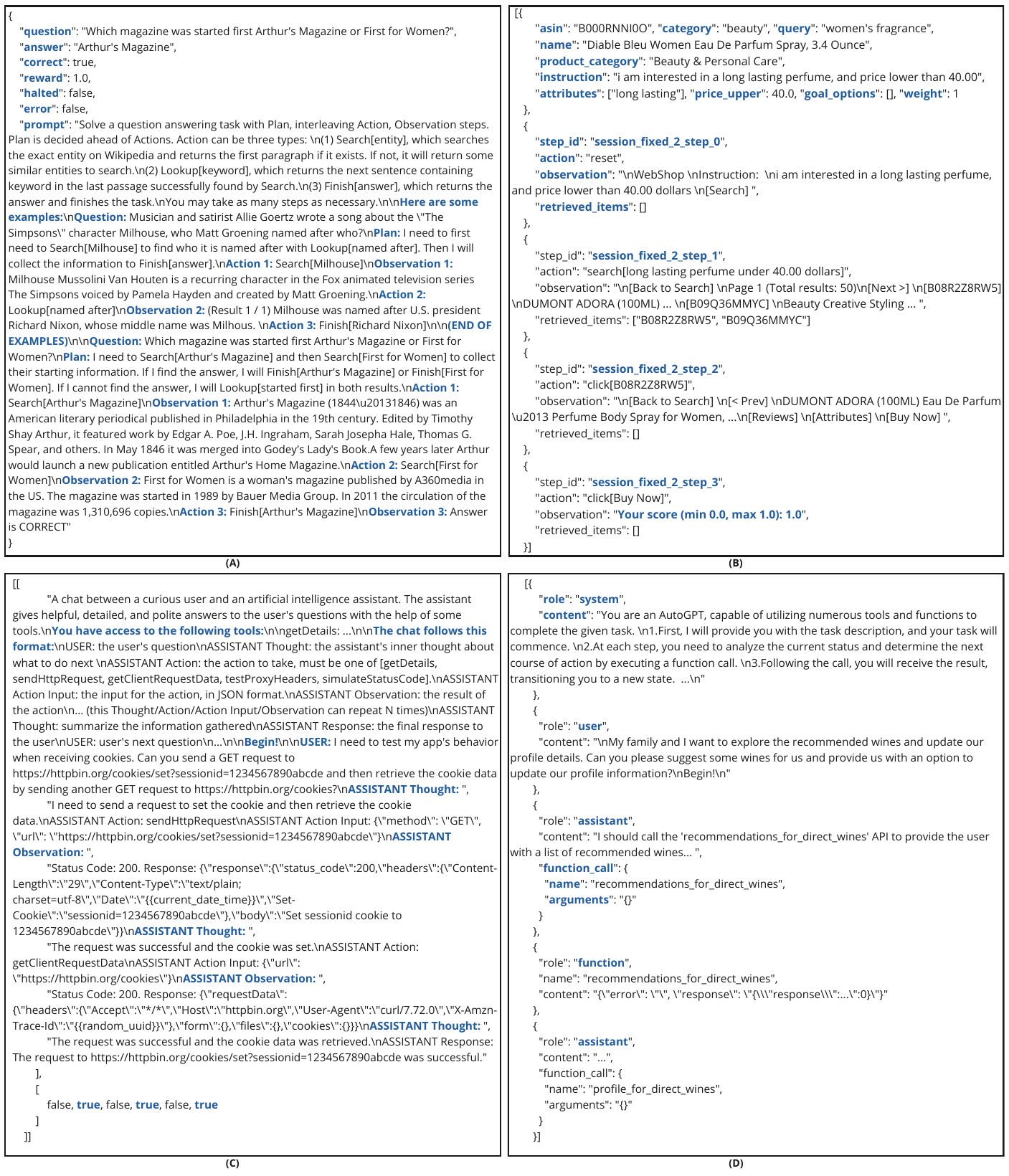}
    \caption{Original trajectories from (A) HotpotQA, (B) Webshop, (C) ToolAlpaca, (D) ToolBench.}
    \label{fig:appendix-raw-trajectory}
\end{figure*}

Figure \ref{fig:appendix-raw-trajectory} shows original trajectories from four environments.

%% file: main.bbl
\begin{thebibliography}{38}
\providecommand{\natexlab}[1]{#1}
\providecommand{\url}[1]{\texttt{#1}}
\expandafter\ifx\csname urlstyle\endcsname\relax
  \providecommand{\doi}[1]{doi: #1}\else
  \providecommand{\doi}{doi: \begingroup \urlstyle{rm}\Url}\fi

\bibitem[Chase(2023)]{langchain23}
Harrison Chase.
\newblock Langchain.
\newblock \url{https://github.com/hwchase17/langchain}, 2023.

\bibitem[Chen et~al.(2023)Chen, Li, Yan, Wang, Gunaratna, Yadav, Tang, Srinivasan, Zhou, Huang, et~al.]{chen2023alpagasus}
Lichang Chen, Shiyang Li, Jun Yan, Hai Wang, Kalpa Gunaratna, Vikas Yadav, Zheng Tang, Vijay Srinivasan, Tianyi Zhou, Heng Huang, et~al.
\newblock Alpagasus: Training a better alpaca with fewer data.
\newblock \emph{arXiv preprint arXiv:2307.08701}, 2023.

\bibitem[Deng et~al.(2023)Deng, Gu, Zheng, Chen, Stevens, Wang, Sun, and Su]{deng2023mind2web}
Xiang Deng, Yu~Gu, Boyuan Zheng, Shijie Chen, Samuel Stevens, Boshi Wang, Huan Sun, and Yu~Su.
\newblock Mind2web: Towards a generalist agent for the web.
\newblock \emph{arXiv preprint arXiv:2306.06070}, 2023.

\bibitem[Dettmers et~al.(2023)Dettmers, Pagnoni, Holtzman, and Zettlemoyer]{dettmers2023qlora}
Tim Dettmers, Artidoro Pagnoni, Ari Holtzman, and Luke Zettlemoyer.
\newblock Qlora: Efficient finetuning of quantized llms, 2023.

\bibitem[Dubois et~al.(2023)Dubois, Li, Taori, Zhang, Gulrajani, Ba, Guestrin, Liang, and Hashimoto]{dubois2023alpacafarm}
Yann Dubois, Xuechen Li, Rohan Taori, Tianyi Zhang, Ishaan Gulrajani, Jimmy Ba, Carlos Guestrin, Percy Liang, and Tatsunori~B. Hashimoto.
\newblock Alpacafarm: A simulation framework for methods that learn from human feedback, 2023.

\bibitem[Gravitas(2023)]{autogpt23}
Significant Gravitas.
\newblock Autogpt.
\newblock \url{https://github.com/Significant-Gravitas/Auto-GPT}, 2023.

\bibitem[Guo et~al.(2024)Guo, Zhang, Liu, Liu, Khalman, Llinares, Rame, Mesnard, Zhao, Piot, et~al.]{guo2024direct}
Shangmin Guo, Biao Zhang, Tianlin Liu, Tianqi Liu, Misha Khalman, Felipe Llinares, Alexandre Rame, Thomas Mesnard, Yao Zhao, Bilal Piot, et~al.
\newblock Direct language model alignment from online ai feedback.
\newblock \emph{arXiv preprint arXiv:2402.04792}, 2024.

\bibitem[Jiang et~al.(2023)Jiang, Sablayrolles, Mensch, Bamford, Chaplot, Casas, Bressand, Lengyel, Lample, Saulnier, et~al.]{jiang2023mistral}
Albert~Q Jiang, Alexandre Sablayrolles, Arthur Mensch, Chris Bamford, Devendra~Singh Chaplot, Diego de~las Casas, Florian Bressand, Gianna Lengyel, Guillaume Lample, Lucile Saulnier, et~al.
\newblock Mistral 7b.
\newblock \emph{arXiv preprint arXiv:2310.06825}, 2023.

\bibitem[Jiang et~al.(2024)Jiang, Sablayrolles, Roux, Mensch, Savary, Bamford, Chaplot, Casas, Hanna, Bressand, et~al.]{jiang2024mixtral}
Albert~Q Jiang, Alexandre Sablayrolles, Antoine Roux, Arthur Mensch, Blanche Savary, Chris Bamford, Devendra~Singh Chaplot, Diego de~las Casas, Emma~Bou Hanna, Florian Bressand, et~al.
\newblock Mixtral of experts.
\newblock \emph{arXiv preprint arXiv:2401.04088}, 2024.

\bibitem[Li et~al.(2023)Li, Song, Yu, Yu, Li, Huang, and Li]{li2023api}
Minghao Li, Feifan Song, Bowen Yu, Haiyang Yu, Zhoujun Li, Fei Huang, and Yongbin Li.
\newblock Api-bank: A benchmark for tool-augmented llms.
\newblock \emph{arXiv preprint arXiv:2304.08244}, 2023.

\bibitem[Liu et~al.(2023{\natexlab{a}})Liu, Yu, Zhang, Xu, Lei, Lai, Gu, Ding, Men, Yang, Zhang, Deng, Zeng, Du, Zhang, Shen, Zhang, Su, Sun, Huang, Dong, and Tang]{liu2023agentbench}
Xiao Liu, Hao Yu, Hanchen Zhang, Yifan Xu, Xuanyu Lei, Hanyu Lai, Yu~Gu, Hangliang Ding, Kaiwen Men, Kejuan Yang, Shudan Zhang, Xiang Deng, Aohan Zeng, Zhengxiao Du, Chenhui Zhang, Sheng Shen, Tianjun Zhang, Yu~Su, Huan Sun, Minlie Huang, Yuxiao Dong, and Jie Tang.
\newblock Agentbench: Evaluating llms as agents, 2023{\natexlab{a}}.

\bibitem[Liu et~al.(2023{\natexlab{b}})Liu, Yao, Zhang, Xue, Heinecke, Murthy, Feng, Chen, Niebles, Arpit, et~al.]{liu2023bolaa}
Zhiwei Liu, Weiran Yao, Jianguo Zhang, Le~Xue, Shelby Heinecke, Rithesh Murthy, Yihao Feng, Zeyuan Chen, Juan~Carlos Niebles, Devansh Arpit, et~al.
\newblock Bolaa: Benchmarking and orchestrating llm-augmented autonomous agents.
\newblock \emph{arXiv preprint arXiv:2308.05960}, 2023{\natexlab{b}}.

\bibitem[Longpre et~al.(2023)Longpre, Hou, Vu, Webson, Chung, Tay, Zhou, Le, Zoph, Wei, et~al.]{longpre2023flan}
Shayne Longpre, Le~Hou, Tu~Vu, Albert Webson, Hyung~Won Chung, Yi~Tay, Denny Zhou, Quoc~V Le, Barret Zoph, Jason Wei, et~al.
\newblock The flan collection: Designing data and methods for effective instruction tuning.
\newblock \emph{arXiv preprint arXiv:2301.13688}, 2023.

\bibitem[Madaan et~al.(2023)Madaan, Tandon, Gupta, Hallinan, Gao, Wiegreffe, Alon, Dziri, Prabhumoye, Yang, et~al.]{madaan2023self}
Aman Madaan, Niket Tandon, Prakhar Gupta, Skyler Hallinan, Luyu Gao, Sarah Wiegreffe, Uri Alon, Nouha Dziri, Shrimai Prabhumoye, Yiming Yang, et~al.
\newblock Self-refine: Iterative refinement with self-feedback.
\newblock \emph{arXiv preprint arXiv:2303.17651}, 2023.

\bibitem[Nijkamp et~al.(2023)Nijkamp, Hayashi, Xiong, Savarese, and Zhou]{nijkamp2023codegen2}
Erik Nijkamp, Hiroaki Hayashi, Caiming Xiong, Silvio Savarese, and Yingbo Zhou.
\newblock Codegen2: Lessons for training llms on programming and natural languages.
\newblock \emph{ICLR}, 2023.

\bibitem[OpenAI(2023)]{openai2023gpt4}
OpenAI.
\newblock Gpt-4 technical report.
\newblock \emph{ArXiv}, 2023.

\bibitem[Qin et~al.(2023)Qin, Liang, Ye, Zhu, Yan, Lu, Lin, Cong, Tang, Qian, et~al.]{qin2023toolllm}
Yujia Qin, Shihao Liang, Yining Ye, Kunlun Zhu, Lan Yan, Yaxi Lu, Yankai Lin, Xin Cong, Xiangru Tang, Bill Qian, et~al.
\newblock Toolllm: Facilitating large language models to master 16000+ real-world apis.
\newblock \emph{arXiv preprint arXiv:2307.16789}, 2023.

\bibitem[Rafailov et~al.(2023)Rafailov, Sharma, Mitchell, Ermon, Manning, and Finn]{rafailov2023direct}
Rafael Rafailov, Archit Sharma, Eric Mitchell, Stefano Ermon, Christopher~D Manning, and Chelsea Finn.
\newblock Direct preference optimization: Your language model is secretly a reward model.
\newblock \emph{arXiv preprint arXiv:2305.18290}, 2023.

\bibitem[Roziere et~al.(2023)Roziere, Gehring, Gloeckle, Sootla, Gat, Tan, Adi, Liu, Remez, Rapin, et~al.]{roziere2023code}
Baptiste Roziere, Jonas Gehring, Fabian Gloeckle, Sten Sootla, Itai Gat, Xiaoqing~Ellen Tan, Yossi Adi, Jingyu Liu, Tal Remez, J{\'e}r{\'e}my Rapin, et~al.
\newblock Code llama: Open foundation models for code.
\newblock \emph{arXiv preprint arXiv:2308.12950}, 2023.

\bibitem[Shinn et~al.(2023)Shinn, Cassano, Gopinath, Narasimhan, and Yao]{shinn2023reflexion}
Noah Shinn, Federico Cassano, Ashwin Gopinath, Karthik~R Narasimhan, and Shunyu Yao.
\newblock Reflexion: Language agents with verbal reinforcement learning.
\newblock In \emph{Thirty-seventh Conference on Neural Information Processing Systems}, 2023.

\bibitem[Shridhar et~al.(2021)Shridhar, Yuan, C\^ot\'e, Bisk, Trischler, and Hausknecht]{ALFWorld20}
Mohit Shridhar, Xingdi Yuan, Marc-Alexandre C\^ot\'e, Yonatan Bisk, Adam Trischler, and Matthew Hausknecht.
\newblock {ALFWorld: Aligning Text and Embodied Environments for Interactive Learning}.
\newblock In \emph{Proceedings of the International Conference on Learning Representations (ICLR)}, 2021.
\newblock URL \url{https://arxiv.org/abs/2010.03768}.

\bibitem[Tang et~al.(2023)Tang, Deng, Lin, Han, Liang, and Sun]{tang2023toolalpaca}
Qiaoyu Tang, Ziliang Deng, Hongyu Lin, Xianpei Han, Qiao Liang, and Le~Sun.
\newblock Toolalpaca: Generalized tool learning for language models with 3000 simulated cases.
\newblock \emph{arXiv preprint arXiv:2306.05301}, 2023.

\bibitem[Team et~al.(2023)Team, Anil, Borgeaud, Wu, Alayrac, Yu, Soricut, Schalkwyk, Dai, Hauth, et~al.]{team2023gemini}
Gemini Team, Rohan Anil, Sebastian Borgeaud, Yonghui Wu, Jean-Baptiste Alayrac, Jiahui Yu, Radu Soricut, Johan Schalkwyk, Andrew~M Dai, Anja Hauth, et~al.
\newblock Gemini: a family of highly capable multimodal models.
\newblock \emph{arXiv preprint arXiv:2312.11805}, 2023.

\bibitem[Team(2023)]{xagent2023}
XAgent Team.
\newblock Xagent: An autonomous agent for complex task solving, 2023.

\bibitem[Touvron et~al.(2023)Touvron, Martin, Stone, Albert, Almahairi, Babaei, Bashlykov, Batra, Bhargava, Bhosale, et~al.]{touvron2023llama}
Hugo Touvron, Louis Martin, Kevin Stone, Peter Albert, Amjad Almahairi, Yasmine Babaei, Nikolay Bashlykov, Soumya Batra, Prajjwal Bhargava, Shruti Bhosale, et~al.
\newblock Llama 2: Open foundation and fine-tuned chat models.
\newblock \emph{arXiv preprint arXiv:2307.09288}, 2023.

\bibitem[Wang et~al.(2023{\natexlab{a}})Wang, Wang, Liu, Chen, Yuan, Peng, and Ji]{wang2023mint}
Xingyao Wang, Zihan Wang, Jiateng Liu, Yangyi Chen, Lifan Yuan, Hao Peng, and Heng Ji.
\newblock Mint: Evaluating llms in multi-turn interaction with tools and language feedback.
\newblock \emph{arXiv preprint arXiv:2309.10691}, 2023{\natexlab{a}}.

\bibitem[Wang et~al.(2023{\natexlab{b}})Wang, Liu, Zhang, Yao, Heinecke, and Yu]{wang2023drdt}
Yu~Wang, Zhiwei Liu, Jianguo Zhang, Weiran Yao, Shelby Heinecke, and Philip~S Yu.
\newblock Drdt: Dynamic reflection with divergent thinking for llm-based sequential recommendation.
\newblock \emph{arXiv preprint arXiv:2312.11336}, 2023{\natexlab{b}}.

\bibitem[Xie et~al.(2023)Xie, Zhou, Cheng, Shi, Weng, Liu, Hua, Zhao, Liu, Liu, et~al.]{xie2023openagents}
Tianbao Xie, Fan Zhou, Zhoujun Cheng, Peng Shi, Luoxuan Weng, Yitao Liu, Toh~Jing Hua, Junning Zhao, Qian Liu, Che Liu, et~al.
\newblock Openagents: An open platform for language agents in the wild.
\newblock \emph{arXiv preprint arXiv:2310.10634}, 2023.

\bibitem[Xu et~al.(2023)Xu, Su, Xing, Mi, Liu, Shi, Hui, Zhou, Liu, Xie, et~al.]{xu2023lemur}
Yiheng Xu, Hongjin Su, Chen Xing, Boyu Mi, Qian Liu, Weijia Shi, Binyuan Hui, Fan Zhou, Yitao Liu, Tianbao Xie, et~al.
\newblock Lemur: Harmonizing natural language and code for language agents.
\newblock \emph{arXiv preprint arXiv:2310.06830}, 2023.

\bibitem[Yang et~al.(2018)Yang, Qi, Zhang, Bengio, Cohen, Salakhutdinov, and Manning]{yang2018hotpotqa}
Zhilin Yang, Peng Qi, Saizheng Zhang, Yoshua Bengio, William~W. Cohen, Ruslan Salakhutdinov, and Christopher~D. Manning.
\newblock {HotpotQA}: A dataset for diverse, explainable multi-hop question answering.
\newblock In \emph{Conference on Empirical Methods in Natural Language Processing ({EMNLP})}, 2018.

\bibitem[Yao et~al.(2022)Yao, Chen, Yang, and Narasimhan]{yao2022webshop}
Shunyu Yao, Howard Chen, John Yang, and Karthik Narasimhan.
\newblock Webshop: Towards scalable real-world web interaction with grounded language agents.
\newblock \emph{Advances in Neural Information Processing Systems}, 35:\penalty0 20744--20757, 2022.

\bibitem[Yao et~al.(2023)Yao, Zhao, Yu, Du, Shafran, Narasimhan, and Cao]{yao2023react}
Shunyu Yao, Jeffrey Zhao, Dian Yu, Nan Du, Izhak Shafran, Karthik Narasimhan, and Yuan Cao.
\newblock {ReAct}: Synergizing reasoning and acting in language models.
\newblock In \emph{International Conference on Learning Representations (ICLR)}, 2023.

\bibitem[Yin et~al.(2023)Yin, Brahman, Ravichander, Chandu, Chang, Choi, and Lin]{yin2023lumos}
Da~Yin, Faeze Brahman, Abhilasha Ravichander, Khyathi Chandu, Kai-Wei Chang, Yejin Choi, and Bill~Yuchen Lin.
\newblock Lumos: Learning agents with unified data, modular design, and open-source llms.
\newblock \emph{arXiv preprint arXiv:2311.05657}, 2023.

\bibitem[Yuan et~al.(2024)Yuan, Pang, Cho, Sukhbaatar, Xu, and Weston]{yuan2024self}
Weizhe Yuan, Richard~Yuanzhe Pang, Kyunghyun Cho, Sainbayar Sukhbaatar, Jing Xu, and Jason Weston.
\newblock Self-rewarding language models.
\newblock \emph{arXiv preprint arXiv:2401.10020}, 2024.

\bibitem[Zeng et~al.(2023)Zeng, Liu, Lu, Wang, Liu, Dong, and Tang]{zeng2023agenttuning}
Aohan Zeng, Mingdao Liu, Rui Lu, Bowen Wang, Xiao Liu, Yuxiao Dong, and Jie Tang.
\newblock Agenttuning: Enabling generalized agent abilities for llms.
\newblock \emph{arXiv preprint arXiv:2310.12823}, 2023.

\bibitem[Zhang et~al.(2023)Zhang, Qian, Liu, Heinecke, Meng, Liu, Yu, Savarese, and Xiong]{zhang2023dialogstudio}
Jianguo Zhang, Kun Qian, Zhiwei Liu, Shelby Heinecke, Rui Meng, Ye~Liu, Zhou Yu, Silvio Savarese, and Caiming Xiong.
\newblock Dialogstudio: Towards richest and most diverse unified dataset collection for conversational ai.
\newblock \emph{arXiv preprint arXiv:2307.10172}, 2023.

\bibitem[Zheng et~al.(2023{\natexlab{a}})Zheng, Chiang, Sheng, Li, Zhuang, Wu, Zhuang, Li, Lin, Xing, et~al.]{zheng2023lmsys}
Lianmin Zheng, Wei-Lin Chiang, Ying Sheng, Tianle Li, Siyuan Zhuang, Zhanghao Wu, Yonghao Zhuang, Zhuohan Li, Zi~Lin, Eric Xing, et~al.
\newblock Lmsys-chat-1m: A large-scale real-world llm conversation dataset.
\newblock \emph{arXiv preprint arXiv:2309.11998}, 2023{\natexlab{a}}.

\bibitem[Zheng et~al.(2023{\natexlab{b}})Zheng, Chiang, Sheng, Zhuang, Wu, Zhuang, Lin, Li, Li, Xing, et~al.]{zheng2023judging}
Lianmin Zheng, Wei-Lin Chiang, Ying Sheng, Siyuan Zhuang, Zhanghao Wu, Yonghao Zhuang, Zi~Lin, Zhuohan Li, Dacheng Li, Eric Xing, et~al.
\newblock Judging llm-as-a-judge with mt-bench and chatbot arena.
\newblock \emph{arXiv preprint arXiv:2306.05685}, 2023{\natexlab{b}}.

\end{thebibliography}
